\documentclass[10pt,twocolumn,letterpaper]{article}

\usepackage{iccv}
\usepackage{times}
\usepackage{epsfig}
\usepackage{graphicx}
\usepackage{color}
\usepackage{amsmath}
\usepackage{amssymb}
\usepackage{multirow}
\usepackage{comment}
\usepackage[british,UKenglish,USenglish,american]{babel}
\usepackage{subcaption}
\usepackage{array}
\graphicspath{{figures/}}

\newcommand{\tabincell}[2]{\begin{tabular}{@{}#1@{}}#2\end{tabular}}
\newcolumntype{x}[1]{>{\centering\arraybackslash\hspace{0pt}}p{#1}}

\usepackage[pagebackref=true,breaklinks=true,letterpaper=true,colorlinks,bookmarks=false]{hyperref}

\iccvfinalcopy 

\def\iccvPaperID{} 

\ificcvfinal\fi
\begin{document}

\title{Deformable Convolutional Networks}

\author{Jifeng Dai\thanks{Equal contribution. \dag This work is done when Haozhi Qi, Yuwen Xiong, Yi Li and Guodong Zhang are interns at Microsoft Research Asia} \quad Haozhi Qi$^{*,\dag}$ \quad Yuwen Xiong$^{*,\dag}$ \quad Yi Li$^{*,\dag}$ \quad Guodong Zhang$^{*,\dag}$ \quad Han Hu \quad Yichen Wei \vspace{8pt}\\
	Microsoft Research Asia\\
	{\tt\small \{jifdai,v-haoq,v-yuxio,v-yii,v-guodzh,hanhu,yichenw\}@microsoft.com}
}

\maketitle

\begin{abstract}

Convolutional neural networks (CNNs) are inherently limited to model geometric transformations due to the fixed geometric structures in their building modules. In this work, we introduce two new modules to enhance the transformation modeling capability of CNNs, namely, deformable convolution and deformable RoI pooling. Both are based on the idea of augmenting the spatial sampling locations in the modules with additional offsets and learning the offsets from the target tasks, without additional supervision. The new modules can readily replace their plain counterparts in existing CNNs and can be easily trained end-to-end by standard back-propagation, giving rise to \emph{deformable convolutional networks}. Extensive experiments validate the performance of our approach. For the first time, we show that learning dense spatial transformation in deep CNNs is effective for sophisticated vision tasks such as object detection and semantic segmentation. The code is released at \url{https://github.com/msracver/Deformable-ConvNets}.

\end{abstract}

\section{Introduction}
\label{sec.introduction}

A key challenge in visual recognition is how to accommodate geometric variations or model geometric transformations in object scale, pose, viewpoint, and part deformation. In general, there are two ways. The first is to build the training datasets with sufficient desired variations. This is usually realized by augmenting the existing data samples, \eg,  by affine transformation. Robust representations can be learned from the data, but usually at the cost of expensive training and complex model parameters. The second is to use transformation-invariant features and algorithms. This category subsumes many well known techniques, such as SIFT (scale invariant feature transform)~\cite{lowe1999object} and sliding window based object detection paradigm.

There are two drawbacks in above ways. First, the geometric transformations are assumed fixed and known. Such prior knowledge is used to augment the data, and design the features and algorithms. This assumption prevents generalization to new tasks possessing unknown geometric transformations, which are not properly modeled. Second, hand-crafted design of invariant features and algorithms could be difficult or infeasible for overly complex transformations, even when they are known.

Recently, convolutional neural networks (CNNs)~\cite{lecun1995convolutional} have achieved significant success for visual recognition tasks, such as image classification~\cite{krizhevsky2012imagenet}, semantic segmentation~\cite{long2015fully}, and object detection~\cite{girshick2014rich}. Nevertheless, they still share the above two drawbacks. Their capability of modeling geometric transformations mostly comes from the extensive data augmentation, the large model capacity, and some simple hand-crafted modules (\eg, max-pooling~\cite{boureau2010pooling} for small translation-invariance).

In short, CNNs are inherently limited to model large, unknown transformations. The limitation originates from the fixed geometric structures of CNN modules: a convolution unit samples the input feature map at fixed locations; a pooling layer reduces the spatial resolution at a fixed ratio; a RoI (region-of-interest) pooling layer separates a RoI into fixed spatial bins, etc. There lacks internal mechanisms to handle the geometric transformations. This causes noticeable problems. \emph{For one example}, the receptive field sizes of all activation units in the same CNN layer are the same. This is undesirable for high level CNN layers that encode the semantics over spatial locations. Because different locations may correspond to objects with different scales or deformation, adaptive determination of scales or receptive field sizes is desirable for visual recognition with fine localization, \eg, semantic segmentation using fully convolutional networks~\cite{long2015fully}. \emph{For another example}, while object detection has seen significant and rapid progress~\cite{girshick2014rich,szegedy2014scalable,girshick2015fast,ren2015faster,redmon2016you,liu2016ssd,dai2016rfcn} recently, all approaches still rely on the primitive bounding box based feature extraction. This is clearly sub-optimal, especially for non-rigid objects.

\begin{figure}[t]
\centering
    \begin{subfigure}{0.24\linewidth}
        \includegraphics[width=\linewidth]{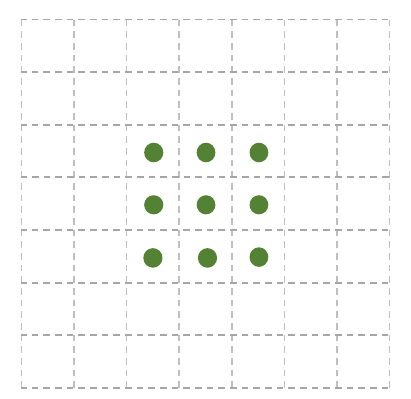}
        \caption{}
    \end{subfigure}
    \begin{subfigure}{0.24\linewidth}
        \includegraphics[width=\linewidth]{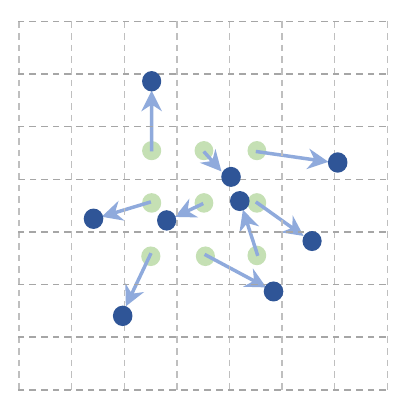}
        \caption{}
    \end{subfigure}
    \begin{subfigure}{0.24\linewidth}
        \includegraphics[width=\linewidth]{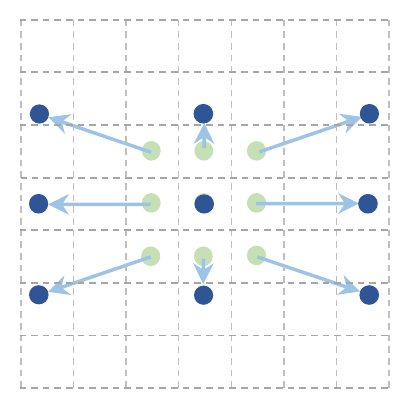}
        \caption{}
    \end{subfigure}
    \begin{subfigure}{0.24\linewidth}
        \includegraphics[width=\linewidth]{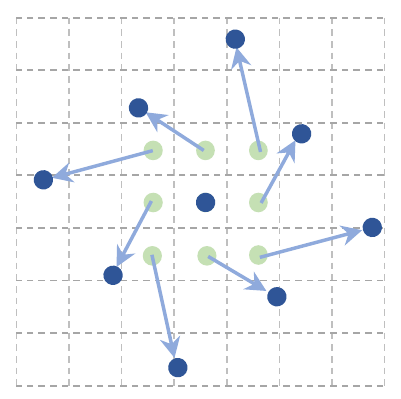}
        \caption{}
    \end{subfigure}
    \vspace{-0.5em}
\caption{Illustration of the sampling locations in $3\times 3$ standard and deformable convolutions. (a) regular sampling grid (green points) of standard convolution. (b) deformed sampling locations (dark blue points) with augmented offsets (light blue arrows) in deformable convolution. (c)(d) are special cases of (b), showing that the deformable convolution generalizes various transformations for scale, (anisotropic) aspect ratio and rotation.}
\label{fig.DC_concept}
\end{figure}

In this work, we introduce two new modules that greatly enhance CNNs' capability of modeling geometric transformations. The first is \emph{deformable convolution}. It adds 2D offsets to the regular grid sampling locations in the standard convolution. It enables free form deformation of the sampling grid. It is illustrated in Figure~\ref{fig.DC_concept}. The offsets are learned from the preceding feature maps, via additional convolutional layers. Thus, the deformation is conditioned on the input features in a local, dense, and adaptive manner.

The second is \emph{deformable RoI pooling}. It adds an offset to each bin position in the regular bin partition of the previous RoI pooling~\cite{girshick2015fast,dai2016rfcn}. Similarly, the offsets are learned from the preceding feature maps and the RoIs, enabling adaptive part localization for objects with different shapes.

Both modules are light weight. They add small amount of parameters and computation for the offset learning. They can readily replace their plain counterparts in deep CNNs and can be easily trained end-to-end with standard backpropagation. The resulting CNNs are called \emph{deformable convolutional networks}, or \emph{deformable ConvNets}.

Our approach shares similar high level spirit with spatial transform networks~\cite{Jaderberg2015} and deformable part models~\cite{pedro2010dpm}. They all have internal transformation parameters and learn such parameters purely from data. A key difference in deformable ConvNets is that they deal with dense spatial transformations in a simple, efficient, deep and end-to-end manner. In Section~\ref{sec.related_work}, we discuss in details the relation of our work to previous works and analyze the superiority of deformable ConvNets.

\section{Deformable Convolutional Networks}

The feature maps and convolution in CNNs are 3D. Both deformable convolution and RoI pooling modules operate on the 2D spatial domain. The operation remains the same across the channel dimension. Without loss of generality, the modules are described in 2D here for notation clarity. Extension to 3D is straightforward.

\begin{figure}
  \centering
  \includegraphics[width=\linewidth]{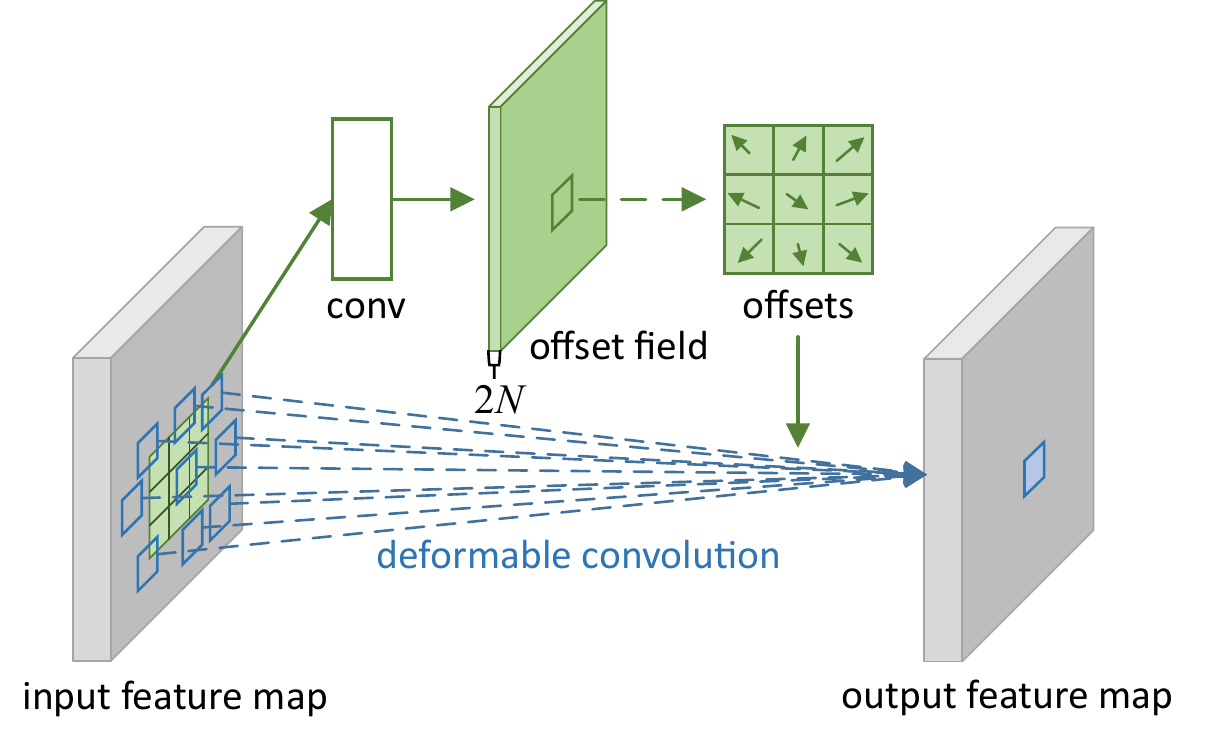}
\caption{Illustration of $3\times 3$ deformable convolution.}
\label{fig.deformable_convolution_structure}
\end{figure}

\subsection{Deformable Convolution}
\label{sec.deformable_convolution}
The 2D convolution consists of two steps: 1) sampling using a regular grid $\mathcal{R}$ over the input feature map $\mathbf{x}$; 2)  summation of sampled values weighted by $\mathbf{w}$. The grid $\mathcal{R}$ defines the receptive field size and dilation. For example, 
\[
\mathcal{R}=\{(-1, -1), (-1, 0), \ldots, (0,1), (1, 1)\}
\]
defines a $3 \times 3$ kernel with dilation $1$.

For each location $\mathbf{p}_0$ on the output feature map $\mathbf{y}$,  we have

\begin{equation}
\mathbf{y}(\mathbf{p}_0)=\sum_{\mathbf{p}_n\in\mathcal{R}}\mathbf{w}(\mathbf{p}_n)\cdot \mathbf{x}(\mathbf{p}_0+\mathbf{p}_n),
\label{eq.standard_conv}
\end{equation}

where $\mathbf{p}_n$ enumerates the locations in $\mathcal{R}$.

In deformable convolution, the regular grid $\mathcal{R}$ is augmented with offsets $\{\Delta \mathbf{p}_n|n=1,...,N\}$, where $N=|\mathcal{R}|$. Eq.~\eqref{eq.standard_conv} becomes

\begin{equation}
\mathbf{y}(\mathbf{p}_0)=\sum_{\mathbf{p}_n\in\mathcal{R}}\mathbf{w}(\mathbf{p}_n)\cdot \mathbf{x}(\mathbf{p}_0+\mathbf{p}_n+\Delta \mathbf{p}_n).
\label{eq.deformable_conv}
\end{equation}

Now, the sampling is on the irregular and offset locations $\mathbf{p}_n+\Delta \mathbf{p}_n$. As the offset $\Delta \mathbf{p}_n$ is typically fractional, Eq.~\eqref{eq.deformable_conv} is implemented via bilinear interpolation as

\begin{equation}
\mathbf{x}(\mathbf{p})=\sum_\mathbf{q} G(\mathbf{q},\mathbf{p})\cdot \mathbf{x}(\mathbf{q}),
\label{eq.bilinear_interpolation}
\end{equation}
where $\mathbf{p}$ denotes an arbitrary (fractional) location ($\mathbf{p}=\mathbf{p}_0+\mathbf{p}_n+\Delta \mathbf{p}_n$ for Eq.~\eqref{eq.deformable_conv}), $\mathbf{q}$ enumerates all integral spatial locations in the feature map $\mathbf{x}$, and $G(\cdot,\cdot)$ is the bilinear interpolation kernel. Note that $G$ is two dimensional. It is separated into two one dimensional kernels as

\begin{equation}
G(\mathbf{q},\mathbf{p})=g(q_x,p_x)\cdot g(q_y,p_y),
\label{eq.bilinear_kernel}
\end{equation}
where $g(a,b)=max(0,1-|a-b|)$. Eq.~\eqref{eq.bilinear_interpolation} is fast to compute as $G(\mathbf{q},\mathbf{p})$ is non-zero only for a few $\mathbf{q}$s.

As illustrated in Figure~\ref{fig.deformable_convolution_structure}, the offsets are obtained by applying a convolutional layer over the same input feature map. The convolution kernel is of the same spatial resolution and dilation as those of the current convolutional layer (\eg, also $3\times 3$ with dilation 1 in Figure~\ref{fig.deformable_convolution_structure}). The output offset fields have the same spatial resolution with the input feature map. The channel dimension $2N$ corresponds to $N$ 2D offsets. During training, both the convolutional kernels for generating the output features and the offsets are learned simultaneously. To learn the offsets, the gradients are back-propagated through the bilinear operations in Eq.~\eqref{eq.bilinear_interpolation} and Eq.~\eqref{eq.bilinear_kernel}. It is detailed in appendix \ref{sec:deformable_conv_backward}.

\begin{figure}
\centering
  \includegraphics[width=\linewidth]{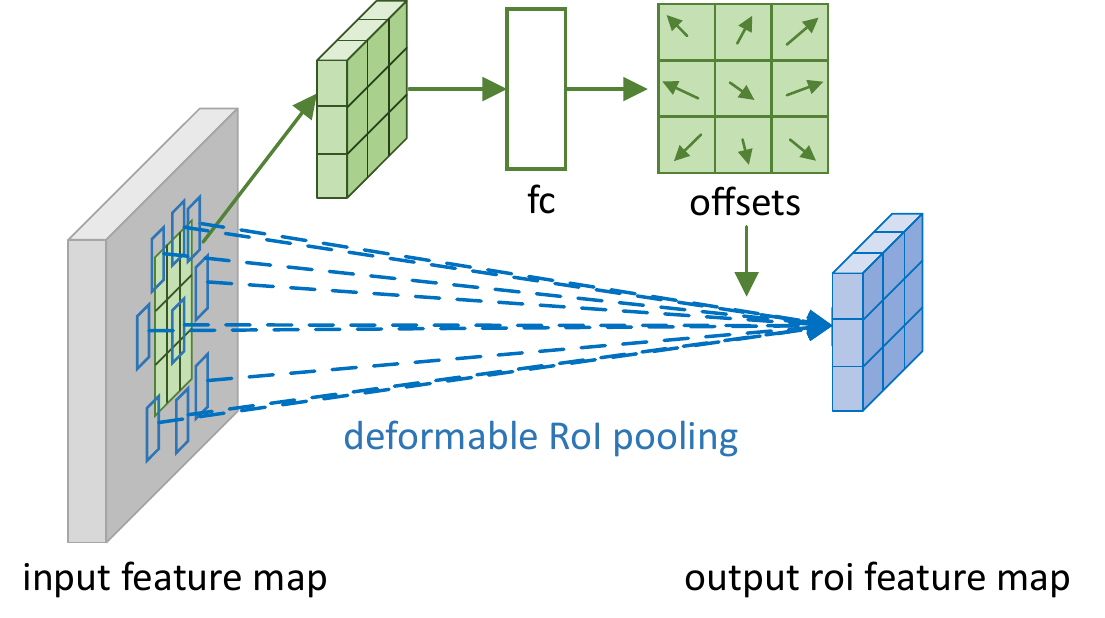}
  \caption{Illustration of $3\times 3$ deformable RoI pooling.}
\label{fig.deformable_roi_pooling_structure}
\end{figure}

\subsection{Deformable RoI Pooling}
\label{sec.deformable_roi_pooling}

\begin{figure}
\centering
  \includegraphics[width=\linewidth]{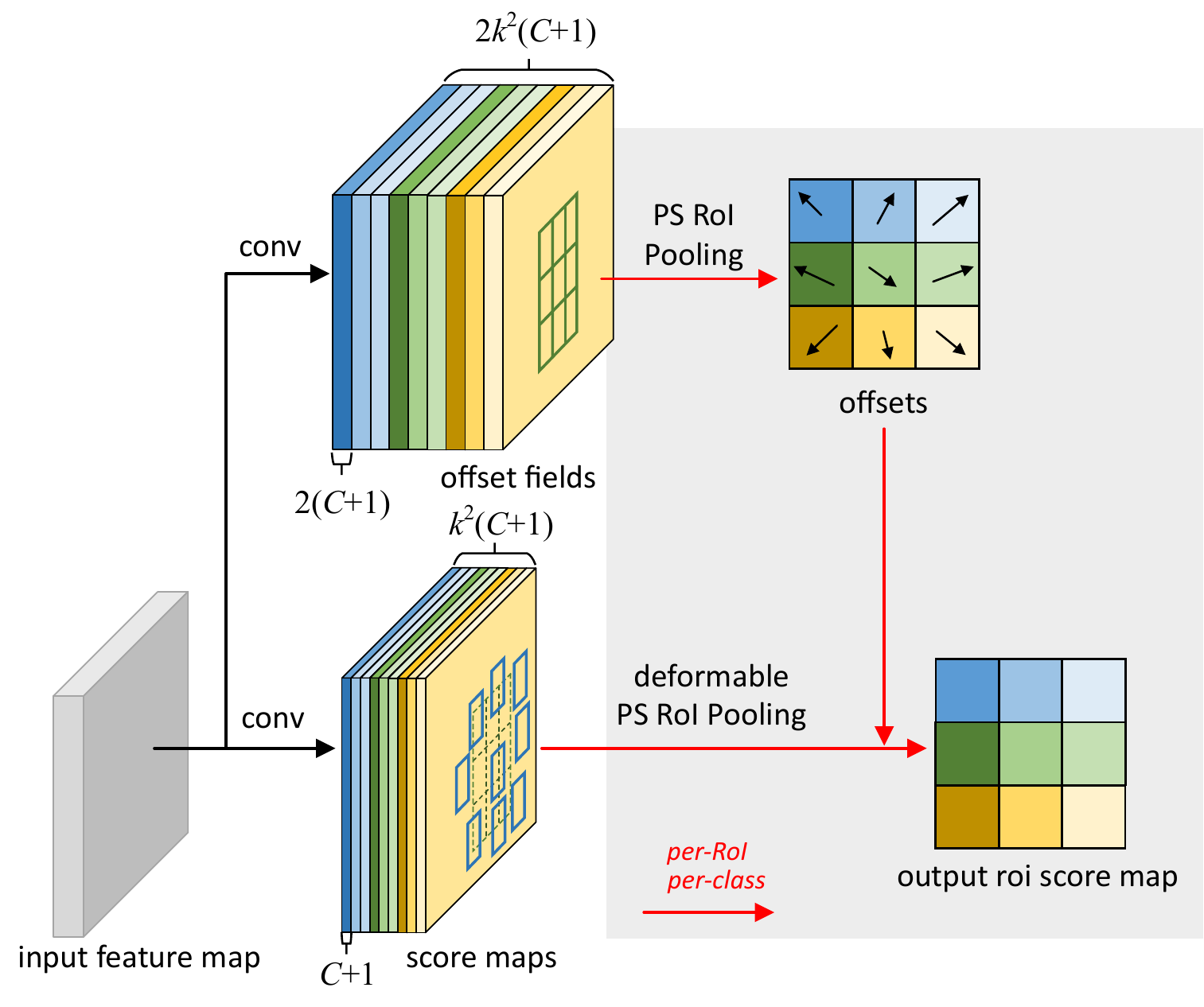}
  \caption{Illustration of $3\times 3$ deformable PS RoI pooling.}
\label{fig.deformable_psroi_pooling_structure}
\end{figure}

RoI pooling is used in all region proposal based object detection methods~\cite{girshick2014rich,girshick2015fast,ren2015faster,dai2016rfcn}. It
converts an input rectangular region of arbitrary size into fixed size features.

\textbf{RoI Pooling~\cite{girshick2015fast}} Given the input feature map $\mathbf{x}$ and a RoI of size $w\times h$ and top-left corner $\mathbf{p}_0$, RoI pooling divides the RoI into $k\times k$ ($k$ is a free parameter) bins and outputs a $k\times k$ feature map $\mathbf{y}$. For $(i,j)$-th bin ($0\le i,j < k$), we have

\begin{equation}
\mathbf{y}(i,j)=\sum_{\mathbf{p}\in bin(i,j)} \mathbf{x}(\mathbf{p}_0+\mathbf{p})/n_{ij},
\label{eq.standard_roi_pooling}
\end{equation}
where $n_{ij}$ is the number of pixels in the bin. The $(i,j)$-th bin spans $\lfloor i \frac{w}{k} \rfloor \le p_x < \lceil (i+1)\frac{w}{k}\rceil$ and $\lfloor j \frac{h}{k}\rfloor \le p_y < \lceil (j+1)\frac{h}{k} \rceil$.

Similarly as in Eq.~\eqref{eq.deformable_conv}, in deformable RoI pooling, offsets $\{\Delta \mathbf{p}_{ij}|0\le i,j < k\}$ are added to the spatial binning positions. Eq.\eqref{eq.standard_roi_pooling} becomes

\begin{equation}
\mathbf{y}(i,j)=\sum_{\mathbf{p}\in bin(i,j)} \mathbf{x}(\mathbf{p}_0+\mathbf{p}+\Delta \mathbf{p}_{ij})/n_{ij}.
\label{eq.deformable_roi_pooling}
\end{equation}
Typically, $\Delta \mathbf{p}_{ij}$ is fractional. Eq.~\eqref{eq.deformable_roi_pooling} is implemented by bilinear interpolation via Eq.~\eqref{eq.bilinear_interpolation} and~\eqref{eq.bilinear_kernel}.

Figure~\ref{fig.deformable_roi_pooling_structure} illustrates how to obtain the offsets. Firstly, RoI pooling (Eq.~\eqref{eq.standard_roi_pooling}) generates the pooled feature maps. From the maps, a fc layer generates the \emph{normalized} offsets $\Delta \widehat{\mathbf{p}}_{ij}$, which are then transformed to the offsets $\Delta \mathbf{p}_{ij}$ in Eq.~\eqref{eq.deformable_roi_pooling} by element-wise product with the RoI's width and height, as $\Delta \mathbf{p}_{ij} = \gamma \cdot \Delta \widehat{\mathbf{p}}_{ij} \circ (w, h)$. Here $\gamma$ is a pre-defined scalar to modulate the magnitude of the offsets. It is empirically set to $\gamma=0.1$. The offset normalization is necessary to make the offset learning invariant to RoI size. The fc layer is learned by back-propagation, as detailed in appendix \ref{sec:deformable_conv_backward}.

\textbf{Position-Sensitive (PS) RoI Pooling~\cite{dai2016rfcn}} It is fully convolutional and different from RoI pooling. Through a conv layer, all the input feature maps are firstly converted to $k^2$ \emph{score maps} for each object class (totally $C+1$ for $C$ object classes), as illustrated in the bottom branch in Figure~\ref{fig.deformable_psroi_pooling_structure}. Without need to distinguish between classes, such score maps are denoted as $\{\mathbf{x}_{i,j}\}$ where $(i,j)$ enumerates all bins. Pooling is performed on these score maps. The output value for $(i,j)$-th bin is obtained by summation from one score map $\mathbf{x}_{i,j}$ corresponding to that bin. In short, the difference from RoI pooling in Eq.\eqref{eq.standard_roi_pooling} is that a general feature map $\mathbf{x}$ is replaced by a specific positive-sensitive score map $\mathbf{x}_{i,j}$. 

In deformable PS RoI pooling, the only change in Eq.~\eqref{eq.deformable_roi_pooling} is that $\mathbf{x}$ is also modified to $\mathbf{x}_{i,j}$. However, the offset learning is different. It follows the ``fully convolutional'' spirit in~\cite{dai2016rfcn}, as illustrated in Figure~\ref{fig.deformable_psroi_pooling_structure}. In the top branch, a conv layer generates the full spatial resolution offset fields. For each RoI (also for each class), PS RoI pooling is applied on such fields to obtain \emph{normalized} offsets $\Delta \widehat{\mathbf{p}}_{ij}$, which are then transformed to the real offsets $\Delta \mathbf{p}_{ij}$ in the same way as in deformable RoI pooling described above.

\subsection{Deformable ConvNets}
\label{sec.deformable_convnets}

Both deformable convolution and RoI pooling modules have the same input and output as their plain versions. Hence, they can readily replace their plain counterparts in existing CNNs. In the training, these added conv and fc layers for offset learning are initialized with zero weights. Their learning rates are set to $\beta$ times ($\beta = 1$ by default, and $\beta = 0.01$ for the fc layer in Faster R-CNN) of the learning rate for the existing layers. They are trained via back propagation through the bilinear interpolation operations in Eq.~\eqref{eq.bilinear_interpolation} and Eq.~\eqref{eq.bilinear_kernel}. The resulting CNNs are called \emph{deformable ConvNets}.

To integrate deformable ConvNets with the state-of-the-art CNN architectures, we note that these architectures consist of two stages. First, a deep fully convolutional network generates feature maps over the whole input image. Second, a shallow task specific network generates results from the feature maps. We elaborate the two steps below.

\textbf{Deformable Convolution for Feature Extraction} We adopt two state-of-the-art architectures for feature extraction: ResNet-101~\cite{he2016deep} and a modifed version of Inception-ResNet~\cite{szegedy2016inception}. Both are pre-trained on ImageNet~\cite{deng2009imagenet} classification dataset.

The original Inception-ResNet is designed for image recognition. It has a feature misalignment issue and problematic for dense prediction tasks. It is modified to fix the alignment problem~\cite{he2016aligned}. The modified version is dubbed as ``Aligned-Inception-ResNet'' and is detailed in appendix \ref{sec:aligned_inception_resnet}.

Both models consist of several convolutional blocks, an average pooling and a 1000-way fc layer for ImageNet classification. The average pooling and the fc layers are removed. A randomly initialized $1 \times 1$ convolution is added at last to reduce the channel dimension to $1024$. As in common practice~\cite{chen2015semantic,dai2016rfcn}, the effective stride in the last convolutional block is reduced from $32$ pixels to $16$ pixels to increase the feature map resolution. Specifically, at the beginning of the last block, stride is changed from $2$ to $1$ (``conv5'' for both ResNet-101 and Aligned-Inception-ResNet). To compensate, the dilation of all the convolution filters in this block (with kernel size $>1$) is changed from $1$ to $2$.

Optionally, \emph{deformable convolution} is applied to the last few convolutional layers (with kernel size $>1$). We experimented with different numbers of such layers and found $3$ as a good trade-off for different tasks, as reported in Table~\ref{table.ablation_usage_deformable_convolution}.

\textbf{Segmentation and Detection Networks} A task specific network is built upon the output feature maps from the feature extraction network mentioned above.

In the below, $C$ denotes the number of object classes.

\emph{DeepLab}~\cite{chen2016deeplab} is a state-of-the-art method for semantic segmentation. It adds a $1 \times 1$ convolutional layer over the feature maps to generates $(C+1)$ maps that represent the per-pixel classification scores. A following softmax layer then outputs the per-pixel probabilities.

\emph{Category-Aware RPN} is almost the same as the region proposal network in~\cite{ren2015faster}, except that the 2-class (object or not) convolutional classifier is replaced by a $(C+1)$-class convolutional classifier. It can be considered as a simplified version of SSD~\cite{liu2016ssd}.

\emph{Faster R-CNN}~\cite{ren2015faster} is the state-of-the-art detector. In our implementation, the RPN branch is added on the top of the conv4 block, following~\cite{ren2015faster}. In the previous practice~\cite{he2016deep,huang2016speed}, the RoI pooling layer is inserted between the conv4 and the conv5 blocks in ResNet-101, leaving 10 layers for each RoI. This design achieves good accuracy but has high per-RoI computation. Instead, we adopt a simplified design as in~\cite{lin2016feature}. The RoI pooling layer is added at last\footnote{The last $1\times 1$ dimension reduction layer is changed to outputs 256-D features.}. On top of the pooled RoI features, two fc layers of dimension $1024$ are added, followed by the bounding box regression and the classification branches. Although such simplification (from 10 layer conv5 block to 2 fc layers) would slightly decrease the accuracy, it still makes a strong enough baseline and is not a concern in this work.

Optionally, the RoI pooling layer can be changed to \emph{deformable RoI pooling}.

\emph{R-FCN}~\cite{dai2016rfcn} is another state-of-the-art detector. It has negligible per-RoI computation cost. We follow the original implementation. Optionally, its RoI pooling layer can be changed to \emph{deformable position-sensitive RoI pooling}.

\section{Understanding Deformable ConvNets}
\label{sec.understanding}

\begin{figure}  
  \begin{subfigure}{.5\linewidth}
  \centering
  \includegraphics[width=\linewidth]{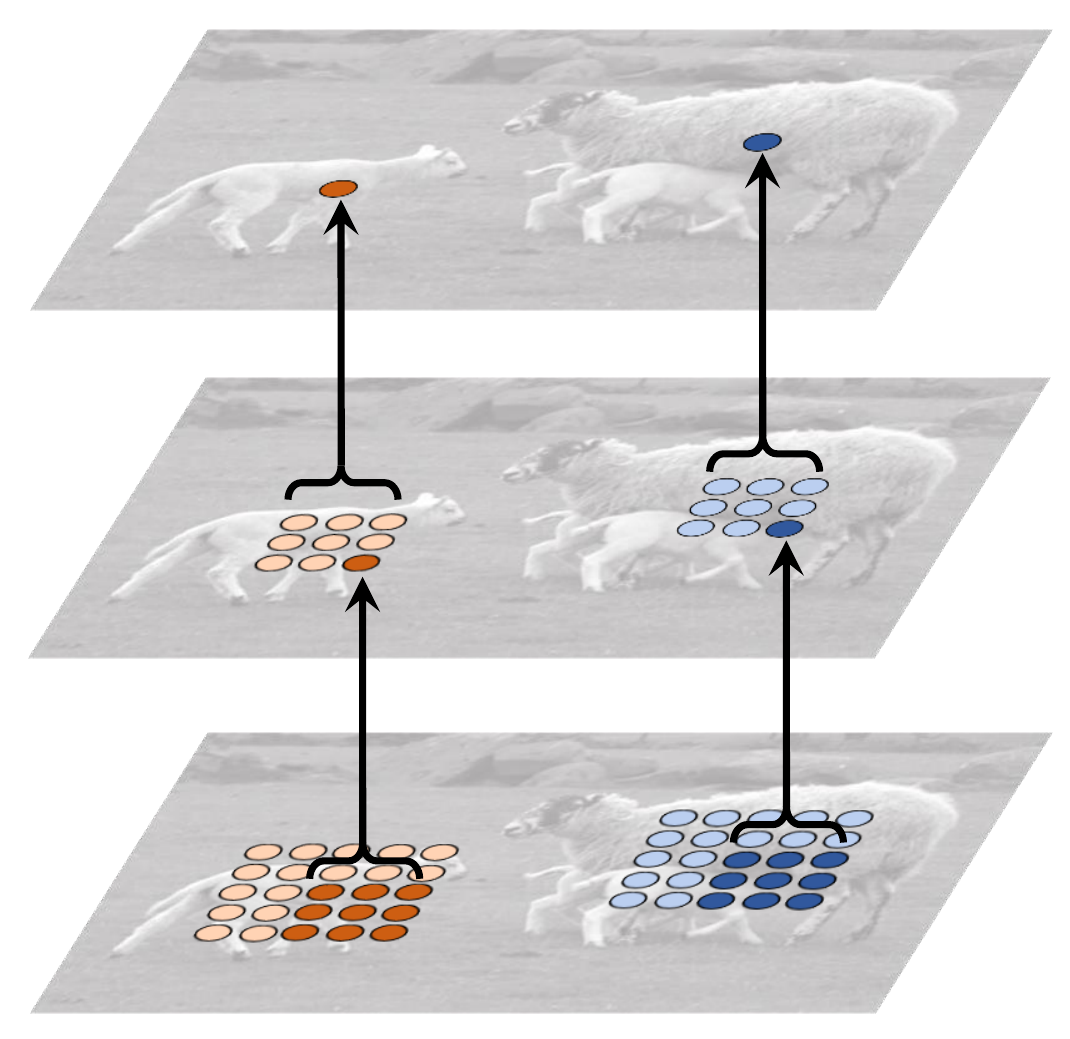}
  \caption{standard convolution}
  \label{fig:dpc_ill_3}
\end{subfigure}%
\begin{subfigure}{.5\linewidth}
  \centering
  \includegraphics[width=\linewidth]{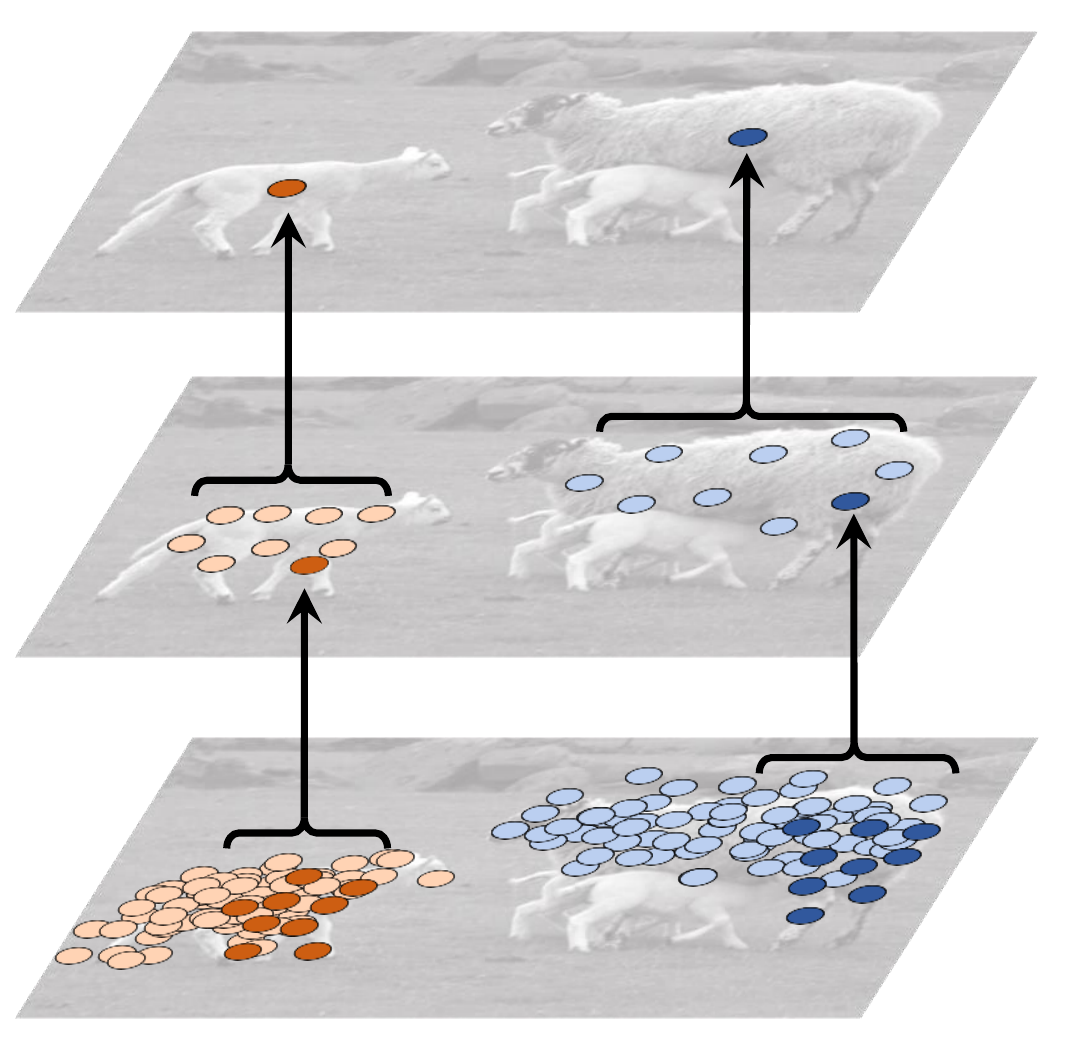}
  \caption{deformable convolution}
  \label{fig:dpc_ill_4}
\end{subfigure}
\caption{Illustration of the fixed receptive field in standard convolution (a) and the adaptive receptive field in deformable convolution (b), using two layers. Top: two activation units on the top feature map, on two objects of different scales and shapes. The activation is from a $3\times 3$ filter. Middle: the sampling locations of the $3\times 3$ filter on the preceding feature map. Another two activation units are highlighted. Bottom: the sampling locations of two levels of $3\times 3$ filters on the preceding feature map. Two sets of locations are highlighted, corresponding to the highlighted units above.}
\label{fig.two_layer_receptive_field_example}
\end{figure}

\begin{figure*}
\centering
\includegraphics[width=\linewidth]{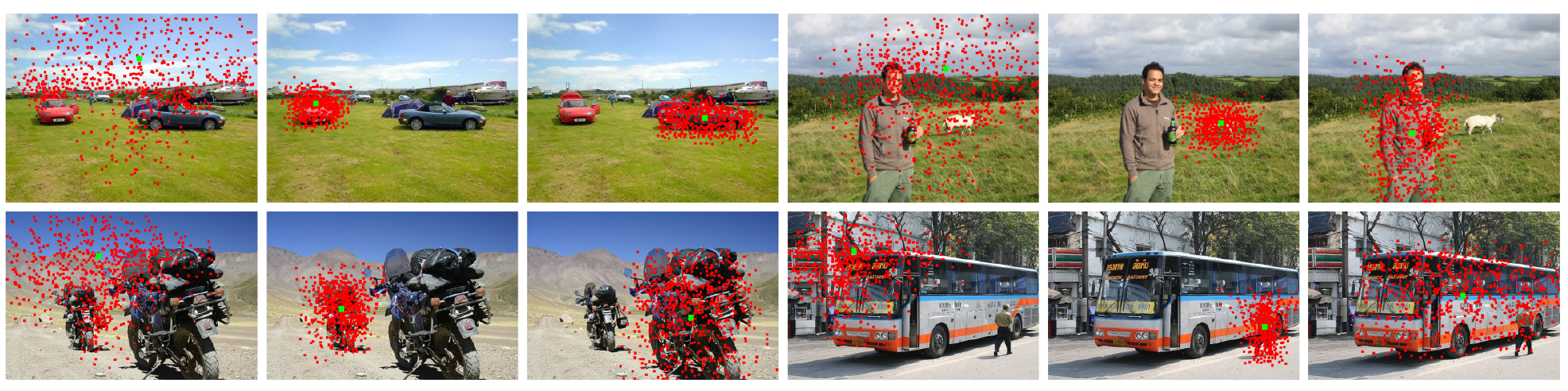}
\caption{Each image triplet shows the sampling locations ($9^3=729$ red points in each image) in three levels of $3\times 3$ deformable filters (see Figure~\ref{fig.two_layer_receptive_field_example} as a reference) for three activation units (green points) on the background (left), a small object (middle), and a large object (right), respectively.}
\label{fig.deform_conv_example}
\end{figure*}

\begin{figure*}
\centering
\includegraphics[width=\linewidth]{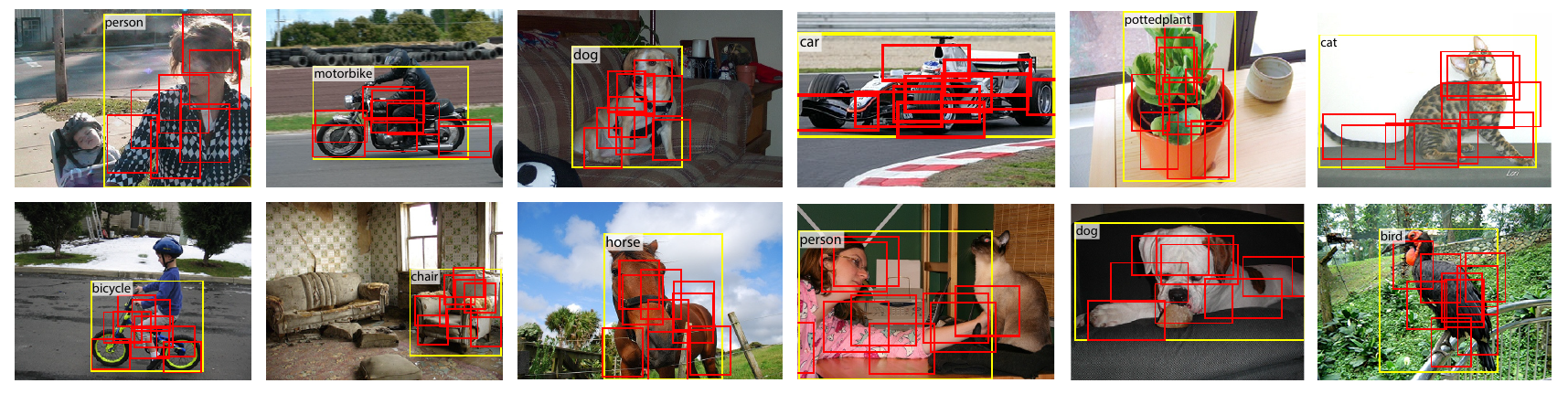}
\caption{Illustration of offset parts in deformable (positive sensitive) RoI pooling in R-FCN~\cite{dai2016rfcn} and $3\times 3$ bins (red) for an input RoI (yellow). Note how the parts are offset to cover the non-rigid objects.}
\label{fig.deform_pool_example}
\end{figure*}

This work is built on the idea of augmenting the spatial sampling locations in convolution and RoI pooling with additional offsets and learning the offsets from target tasks. 

When the deformable convolution are stacked, the effect of composited deformation is profound. This is exemplified in Figure~\ref{fig.two_layer_receptive_field_example}. The receptive field and the sampling locations in the standard convolution are fixed all over the top feature map (left). They are adaptively adjusted according to the objects' scale and shape in deformable convolution (right). More examples are shown in Figure~\ref{fig.deform_conv_example}. Table~\ref{table.deformable_convolution_stat} provides quantitative evidence of such adaptive deformation.

The effect of deformable RoI pooling is similar, as illustrated in Figure~\ref{fig.deform_pool_example}. The regularity of the grid structure in standard RoI pooling no longer holds. Instead, parts deviate from the RoI bins and move onto the nearby object foreground regions. The localization capability is enhanced, especially for non-rigid objects.

\subsection{In Context of Related Works}
\label{sec.related_work}

Our work is related to previous works in different aspects. We discuss the relations and differences in details.

\textbf{Spatial Transform Networks (STN)~\cite{Jaderberg2015}} It is the first work to learn spatial transformation from data in a deep learning framework. \emph{It warps the feature map via a global parametric transformation} such as affine transformation. Such warping is expensive and learning the transformation parameters is known difficult. STN has shown successes in small scale image classification problems. The inverse STN method~\cite{lin2016inverse} replaces the expensive feature warping by efficient transformation parameter propagation.

The offset learning in deformable convolution can be considered as an extremely light-weight spatial transformer in STN~\cite{Jaderberg2015}. However, \emph{deformable convolution does not adopt a global parametric transformation and feature warping. Instead, it samples the feature map in a local and dense manner}. To generate new feature maps, it has a weighted summation step, which is absent in STN.

Deformable convolution is easy to integrate into any CNN architectures. Its training is easy. It is shown effective for complex vision tasks that require dense (\eg, semantic segmentation) or semi-dense (\eg, object detection) predictions. These tasks are difficult (if not infeasible) for STN~\cite{Jaderberg2015,lin2016inverse}.

\textbf{Active Convolution~\cite{jeon_cvpr2017active}} This work is contemporary. It also augments the sampling locations in the convolution with offsets and learns the offsets via back-propagation end-to-end. It is shown effective on image classification tasks. 

Two crucial differences from deformable convolution make this work less general and adaptive. First, it shares the offsets all over the different spatial locations. Second, the offsets are \emph{static model parameters} that are learnt per task or per training. In contrast, the offsets in deformable convolution are \emph{dynamic model outputs}  that vary per image location. They model the dense spatial transformations in the images and are effective for (semi-)dense prediction tasks such as object detection and semantic segmentation.

\textbf{Effective Receptive Field~\cite{luo2017understanding}}  It finds that not all pixels in a receptive field contribute equally to an output response. The pixels near the center have much larger impact. The effective receptive field only occupies a small fraction of the theoretical receptive field and has a Gaussian distribution. Although the theoretical receptive field size increases linearly with the number of convolutional layers, a surprising result is that, the effective receptive field size increases linearly with the \emph{square root} of the number, therefore, \emph{at a much slower rate} than what we would expect.

This finding indicates that even the top layer's unit in deep CNNs may not have large enough receptive field. This partially explains why atrous convolution~\cite{Holschneider89} is widely used in vision tasks (see below). It indicates the needs of adaptive receptive field learning.

Deformable convolution is capable of learning receptive fields adaptively, as shown in Figure~\ref{fig.two_layer_receptive_field_example},~\ref{fig.deform_conv_example} and Table~\ref{table.deformable_convolution_stat}.

\textbf{Atrous convolution~\cite{Holschneider89}} It increases a normal filter's stride to be larger than $1$ and keeps the original weights at sparsified sampling locations. This increases the receptive field size and retains the same complexity in parameters and computation. It has been widely used for semantic segmentation~\cite{long2015fully,chen2016deeplab,fisher2016dilated} (also called dilated convolution in~\cite{fisher2016dilated}), object detection~\cite{dai2016rfcn}, and image classification~\cite{yu2017dilated}.

Deformable convolution is a generalization of atrous convolution, as easily seen in Figure~\ref{fig.DC_concept} (c). Extensive comparison to atrous convolution is presented in Table~\ref{table.ablation_VOC}.

\setlength{\tabcolsep}{2pt}
\renewcommand{\arraystretch}{1.2}
\begin{table*}[t]
\centering
\small
\begin{tabular}{l|c|c|c|c|c|c|c|c}
\hline
\multirow{2}{*}{\tabincell{c}{usage of deformable \\ convolution ($\#$ layers)} } & \multicolumn{2}{c|}{DeepLab}  & \multicolumn{2}{c|}{class-aware RPN} & \multicolumn{2}{c|}{Faster R-CNN} & \multicolumn{2}{c}{R-FCN} \\
\cline{2-9}
& \scriptsize{mIoU@V (\%)} & \scriptsize{mIoU@C (\%)} & \scriptsize{mAP@0.5 (\%)} & \scriptsize{mAP@0.7 (\%)} & \scriptsize{mAP@0.5 (\%)}  & \scriptsize{mAP@0.7 (\%)} & \scriptsize{mAP@0.5 (\%)} & \scriptsize{mAP@0.7 (\%)} \\ 
\hline\hline
none (0, baseline) & 69.7 & 70.4 & $68.0$ & $44.9$ & 78.1 & 62.1 & 80.0 & 61.8 \\
\hline
res5c (1)      & 73.9 & 73.5 & $73.5$ & $54.4$ & 78.6 & 63.8 & 80.6 & 63.0 \\
res5b,c (2)    & 74.8 &  74.4 & $74.3$ & $56.3$ & 78.5 & 63.3 & 81.0 & 63.8 \\
res5a,b,c (3, default) & $\mathbf{75.2}$ & $\mathbf{75.2}$ & $74.5$ & $57.2$ & 78.6 & 63.3 & 81.4 & 64.7 \\
res5 \& res4b22,b21,b20 (6) & 74.8 & 75.1 & $\mathbf{74.6}$ & $\mathbf{57.7}$ & $\mathbf{78.7}$ & $\mathbf{64.0}$ & $\mathbf{81.5}$ & $\mathbf{65.4}$ \\
\hline
\end{tabular}
\caption{Results of using deformable convolution in the last $1$, $2$, $3$, and $6$ convolutional layers (of $3 \times 3$ filter) in ResNet-101 feature extraction network. For \emph{class-aware RPN}, \emph{Faster R-CNN}, and \emph{R-FCN}, we report result on VOC 2007 test.}
\label{table.ablation_usage_deformable_convolution}
\end{table*}

\textbf{Deformable Part Models (DPM)~\cite{pedro2010dpm}} Deformable RoI pooling is similar to DPM because both methods learn the spatial deformation of object parts to maximize the classification score. Deformable RoI pooling is simpler since no spatial relations between the parts are considered.

DPM is a shallow model and has limited capability of modeling deformation. While its inference algorithm can be converted to CNNs~\cite{girshick2014deformable} by treating the distance transform as a special pooling operation, its training is not end-to-end and involves heuristic choices such as selection of components and part sizes. In contrast, deformable ConvNets are deep and perform end-to-end training. When multiple deformable modules are stacked, the capability of modeling deformation becomes stronger.

\textbf{DeepID-Net~\cite{ouyang2015deepid}} It introduces a deformation constrained pooling layer which also considers part deformation for object detection. It therefore shares a similar spirit with deformable RoI pooling, but is much more complex. This work is highly engineered and based on RCNN~\cite{girshick2014rich}. It is unclear how to adapt it to the recent state-of-the-art object detection methods~\cite{ren2015faster,dai2016rfcn} in an end-to-end manner.

\textbf{Spatial manipulation in RoI pooling} Spatial pyramid pooling~\cite{lazebnik2006beyond} uses hand crafted pooling regions over scales. It is the predominant approach in computer vision and also used in deep learning based object detection~\cite{he2014spatial,girshick2015fast}.

Learning the spatial layout of pooling regions has received little study. The work in~\cite{jia2012beyond} learns a sparse subset of pooling regions from a large over-complete set. The large set is hand engineered and the learning is not end-to-end.

Deformable RoI pooling is the first to learn pooling regions end-to-end in CNNs. While the regions are of the same size currently, extension to multiple sizes as in spatial pyramid pooling~\cite{lazebnik2006beyond} is straightforward.

\textbf{Transformation invariant features and their learning} There have been tremendous efforts on designing transformation invariant features. Notable examples include scale invariant feature transform (SIFT)~\cite{lowe1999object} and ORB~\cite{rublee2011orb} (O for orientation). There is a large body of such works in the context of CNNs. The invariance and equivalence of CNN representations to image transformations are studied in~\cite{lenc2015understanding}. Some works learn invariant CNN representations with respect to different types of transformations such as~\cite{sohn2012invariant}, scattering networks~\cite{bruna2013scattering}, convolutional jungles~\cite{laptev2015transformation}, and TI-pooling ~\cite{laptev2016tipooling}. Some works are devoted for specific transformations such as symmetry~\cite{gens2014deepsymmetry,dieleman2016cyclic}, scale~\cite{kanazawa2014scale}, and rotation~\cite{worrall2016harmonic}.

As analyzed in Section~\ref{sec.introduction}, in these works the transformations are known a priori. The knowledge (such as parameterization) is used to hand craft the structure of feature extraction algorithm, either fixed in such as SIFT, or with learnable parameters such as those based on CNNs. They cannot handle unknown transformations in the new tasks.

In contrast, our deformable modules generalize various transformations (see Figure~\ref{fig.DC_concept}). The transformation invariance is learned from the target task.

\textbf{Dynamic Filter~\cite{bert2016dynamic}} Similar to deformable convolution, the dynamic filters are also conditioned on the input features and change over samples. Differently, only the filter weights are learned, not the sampling locations like ours. This work is applied for video and stereo prediction.

\textbf{Combination of low level filters} Gaussian filters and its smooth derivatives~\cite{Koenderink87} are widely used to extract low level image structures such as corners, edges, T-junctions, etc. Under certain conditions, such filters form a set of basis and their linear combination forms new filters within the same group of geometric transformations, such as multiple orientations in \emph{Steerable Filters}~\cite{freeman1991steerable} and multiple scales in~\cite{perona1995deformable}. We note that although the term \emph{deformable kernels} is used in ~\cite{perona1995deformable}, its meaning is different from ours in this work.

Most CNNs learn all their convolution filters from scratch. The recent work~\cite{jacobsen2016structured} shows that it could be unnecessary. It replaces the free form filters by weighted combination of low level filters (Gaussian derivatives up to 4-th order) and learns the weight coefficients. The regularization over the filter function space is shown to improve the generalization ability when training data are small.

Above works are related to ours in that, when multiple filters, especially with different scales, are combined, the resulting filter could have complex weights and resemble our deformable convolution filter. However, deformable convolution learns sampling locations instead of filter weights.

\setlength{\tabcolsep}{4pt}
\renewcommand{\arraystretch}{1.2}
\begin{table}[t]
\centering
\small
\begin{tabular}{l|c|c|c|c}
\hline
\multirow{2}{*}{layer} &  small &  medium & large &  background \\ 
\cline{2-5}
& \multicolumn{4}{c}{ mean $\pm$  std} \\
\hline\hline
res5c & 5.3 $\pm$ 3.3 & 5.8 $\pm$ 3.5 & 8.4 $\pm$ 4.5 & 6.2 $\pm$ 3.0 \\
res5b & 2.5 $\pm$ 1.3 & 3.1 $\pm$ 1.5 & 5.1 $\pm$ 2.5 & 3.2 $\pm$ 1.2 \\
res5a & 2.2 $\pm$ 1.2 & 2.9 $\pm$ 1.3 & 4.2 $\pm$ 1.6 & 3.1 $\pm$ 1.1 \\
\hline
\end{tabular}
\caption{Statistics of effective dilation values of deformable convolutional filters on three layers and four categories. Similar as in COCO~\cite{lin2014coco}, we divide the objects into three categories equally according to the bounding box area. Small: area $< 96^2$ pixels; medium: $96^2 < $ area $ < 224^2$; large: area $>224^2$ pixels.}
\label{table.deformable_convolution_stat}
\end{table}

\setlength{\tabcolsep}{8pt}
\renewcommand{\arraystretch}{1.2}
\begin{table*}[t]
\centering
\small
\begin{tabular}{l|c|c|c|c}
\hline
deformation modules & \tabincell{c}{DeepLab \\ {\footnotesize mIoU@V / @C}} & \tabincell{c}{class-aware RPN \\ {\footnotesize mAP@0.5 / @0.7}} & \tabincell{c}{Faster R-CNN \\ {\footnotesize mAP@0.5 / @0.7}} & \tabincell{c}{R-FCN \\ {\footnotesize mAP@0.5 / @0.7}} \\
\hline
\hline
atrous convolution (2,2,2) (default) &   69.7 / 70.4  & 68.0 / 44.9 &  78.1 / 62.1  &  80.0 / 61.8  \\
atrous convolution (4,4,4) &   73.1 / 71.9  & 72.8 / 53.1 &  78.6 / 63.1  &  80.5 / 63.0  \\
atrous convolution (6,6,6) &   73.6 / 72.7 & 73.6 / 55.2 &  78.5 / 62.3  & 80.2 / 63.5    \\
atrous convolution (8,8,8) &   73.2 / 72.4 & 73.2 / 55.1 &  77.8 / 61.8  &   80.3 / 63.2   \\
\hline
deformable convolution      &   $\mathbf{75.3}$ / $\mathbf{75.2}$  & $\mathbf{74.5}$ / $\mathbf{57.2}$ &  78.6 / 63.3 & 81.4 / 64.7  \\
\hline
deformable RoI pooling    &  N.A   &  N.A         &  78.3 / 66.6   &  81.2 / 65.0  \\
deformable convolution \& RoI pooling    &  N.A   &  N.A   & $\mathbf{79.3}$ / $\mathbf{66.9}$   & $\mathbf{82.6}$ / $\mathbf{68.5}$  \\
\hline
\end{tabular}
\caption{Evaluation of our deformable modules and atrous convolution, using ResNet-101.}
\label{table.ablation_VOC}
\end{table*}

\setlength{\tabcolsep}{4.0pt}
\renewcommand{\arraystretch}{1.2}
\begin{table}[t]
\centering
\small
\begin{tabular}{l|c|c|c}
\hline
method & \# params & \renewcommand{\arraystretch}{1.0} \tabincell{c}{net. forward \\ (sec)} & \renewcommand{\arraystretch}{1.0} \tabincell{c}{runtime \\ (sec)} \\
\hline
\hline
DeepLab@C       & $46.0$ M & 0.610 & 0.650 \\
\textbf{Ours}   & $46.1$ M & 0.656 & 0.696 \\
\hline
DeepLab@V       & $46.0$ M &  0.084 & 0.094 \\
\textbf{Ours}   & $46.1$ M & 0.088 & 0.098 \\
\hline
class-aware RPN & $46.0$ M & 0.142 & 0.323 \\
\textbf{Ours}   & $46.1$ M & 0.152 & 0.334 \\
\hline
Faster R-CNN    & $58.3$ M & 0.147 & 0.190 \\
\textbf{Ours}   & $59.9$ M &  0.192 & 0.234 \\
\hline
R-FCN           & $47.1$ M & 0.143 & 0.170  \\
\textbf{Ours}   & $49.5$ M & 0.169 & 0.193  \\
\hline
\end{tabular}
\caption{Model complexity and runtime comparison of deformable ConvNets and the plain counterparts, using ResNet-101. The overall runtime in the last column includes image resizing, network forward, and post-processing (\eg, NMS for object detection). Runtime is counted on a workstation with Intel E5-2650 v2 CPU and Nvidia K40 GPU.}
\label{table.complexity_memory_runtime}
\end{table}

\setlength{\tabcolsep}{3pt}
\renewcommand{\arraystretch}{1.2}
\begin{table*}[t]
\centering
\small
\begin{tabular}{l|c|x{0.8cm}x{0.8cm}|c|c|c|c|c}
\hline
method & \tabincell{c}{backbone \\ architecture } & M & B & \footnotesize mAP@[0.5:0.95] & \scriptsize mAP$^r$@0.5 & \renewcommand{\arraystretch}{1.1} \scriptsize \tabincell{c}{mAP@[0.5:0.95] \\ (small)} & \renewcommand{\arraystretch}{1.1} \scriptsize \tabincell{c}{mAP@[0.5:0.95] \\ (mid)} & \renewcommand{\arraystretch}{1.1} \scriptsize \tabincell{c}{mAP@[0.5:0.95] \\ (large)}\\
\hline\hline
class-aware RPN & \multirow{2}{*}{\footnotesize ResNet-101} & & & 23.2 & 42.6 & 6.9 & 27.1 & 35.1 \\
\textbf{Ours}  & & & & $\mathbf{25.8}$ & $\mathbf{45.9}$ & $\mathbf{7.2}$ & $\mathbf{28.3}$ & $\mathbf{40.7}$ \\
\hline
Faster RCNN & \multirow{2}{*}{\footnotesize ResNet-101} & & & 29.4 & 48.0 & 9.0 & 30.5 & 47.1\\
\textbf{Ours} &  & & & $\mathbf{33.1}$ & $\mathbf{50.3}$ & $\mathbf{11.6}$ & $\mathbf{34.9}$ & $\mathbf{51.2}$ \\
\hline
R-FCN & \multirow{2}{*}{\footnotesize ResNet-101} & & & 30.8 & 52.6 & 11.8 & 33.9 & 44.8 \\
\textbf{Ours} &  & & & $\mathbf{34.5}$ & $\mathbf{55.0}$ & $\mathbf{14.0}$ & $\mathbf{37.7}$ & $\mathbf{50.3}$ \\
\hline
Faster RCNN & \multirow{2}{*}{\tabincell{c}{\footnotesize Aligned-Inception-ResNet}} & & & 30.8 & 49.6 & 9.6 & 32.5 & 49.0 \\
\textbf{Ours} & & & & $\mathbf{34.1}$ & $\mathbf{51.1}$ & $\mathbf{12.2}$ & $\mathbf{36.5}$ & $\mathbf{52.4}$ \\
\hline
R-FCN & \multirow{2}{*}{\tabincell{c}{\footnotesize Aligned-Inception-ResNet}} & & & 32.9 & 54.5 & 12.5 & 36.3 & 48.3 \\
\textbf{Ours} &  & & & $\mathbf{36.1}$ & $\mathbf{56.7}$ & $\mathbf{14.8}$ & $\mathbf{39.8}$ & $\mathbf{52.2}$ \\
\hline
R-FCN &  & \checkmark & & 34.5 & 55.0 & 16.8 & 37.3 & 48.3 \\
\textbf{Ours} & \multirow{2}{*}{\tabincell{c}{\footnotesize Aligned-Inception-ResNet}}  & \checkmark & & 37.1 & 57.3 & 18.8 & 39.7 & 52.3 \\
R-FCN &  & \checkmark & \checkmark & 35.5 & 55.6 & 17.8 & 38.4 & 49.3 \\
\textbf{Ours} &  & \checkmark & \checkmark & $\mathbf{37.5}$ & $\mathbf{58.0}$ & $\mathbf{19.4}$ & $\mathbf{40.1}$ & $\mathbf{52.5}$ \\
\hline
\end{tabular}
\caption{Object detection results of deformable ConvNets v.s. plain ConvNets on COCO test-dev set. M denotes multi-scale testing, and B denotes iterative bounding box average in the table.}
\label{table.detection_on_coco}
\end{table*}

\section{Experiments}

\subsection{Experiment Setup and Implementation}
\label{sec.experiment_setup}

\textbf{Semantic Segmentation}
We use \emph{PASCAL VOC}~\cite{everingham2010pascal} and \emph{CityScapes}~\cite{cordts2016cityscapes}. For \emph{PASCAL VOC}, there are $20$ semantic categories. Following the protocols in~\cite{hariharan2014simultaneous,long2015fully,chen2015semantic}, we use VOC 2012 dataset and the additional mask annotations in \cite{hariharan2011semantic}. The training set includes $10,582$ images. Evaluation is performed on $1,449$ images in the validation set. For \emph{CityScapes}, following the protocols in~\cite{chen2016deeplab}, training and evaluation are performed on $2,975$ images in the train set and $500$ images in the validation set, respectively. There are $19$ semantic categories plus a background category.

For evaluation, we use the mean intersection-over-union (mIoU) metric defined over image pixels, following the standard protocols~\cite{everingham2010pascal,cordts2016cityscapes}. We use mIoU@V and mIoU@C for PASCAl VOC and Cityscapes, respectively.

In training and inference, the images are resized to have a shorter side of $360$ pixels for PASCAL VOC and $1,024$ pixels for Cityscapes. In SGD training, one image is randomly sampled in each mini-batch. A total of 30k and 45k iterations are performed for PASCAL VOC and Cityscapes, respectively, with 8 GPUs and one mini-batch on each. The learning rates are $10^{-3}$ and $10^{-4}$ in the first $\frac{2}{3}$ and the last $\frac{1}{3}$ iterations, respectively.

\textbf{Object Detection}
We use \emph{PASCAL VOC} and \emph{COCO}~\cite{lin2014coco} datasets. For \emph{PASCAL VOC}, following the protocol in~\cite{girshick2015fast}, training is performed on the union of VOC 2007 trainval and VOC 2012 trainval. Evaluation is on VOC 2007 test. For \emph{COCO}, following the standard protocol~\cite{lin2014coco}, training and evaluation are performed on the 120k images in the trainval and the 20k images in the test-dev, respectively.

For evaluation, we use the standard mean average precision (mAP) scores~\cite{everingham2010pascal,lin2014coco}. For PASCAL VOC, we report mAP scores using IoU thresholds at 0.5 and 0.7. For COCO, we use the standard COCO metric of mAP@[0.5:0.95], as well as mAP@0.5.

In training and inference, the images are resized to have a shorter side of 600 pixels. In SGD training, one image is randomly sampled in each mini-batch. For \emph{class-aware RPN}, 256 RoIs are sampled from the image. For \emph{Faster R-CNN} and \emph{R-FCN}, 256 and 128 RoIs are sampled for the region proposal and the object detection networks, respectively. $7\times 7$ bins are adopted in RoI pooling. To facilitate the ablation experiments on VOC, we follow~\cite{lin2016feature} and utilize pre-trained and fixed RPN proposals for the training of Faster R-CNN and R-FCN, without feature sharing between the region proposal and the object detection networks. The RPN network is trained separately as in the first stage of the procedure in~\cite{ren2015faster}. For COCO, joint training as in~\cite{ren2016faster} is performed and feature sharing is enabled for training. A total of 30k and 240k iterations are performed for PASCAL VOC and COCO, respectively, on 8 GPUs. The learning rates are set as $10^{-3}$ and $10^{-4}$ in the first $\frac{2}{3}$ and the last $\frac{1}{3}$ iterations, respectively.

\subsection{Ablation Study}
\label{sec.experiment_ablation}

Extensive ablation studies are performed to validate the efficacy and efficiency of our approach.

\textbf{Deformable Convolution} Table~\ref{table.ablation_usage_deformable_convolution} evaluates the effect of deformable convolution using ResNet-101 feature extraction network. Accuracy steadily improves when more deformable convolution layers are used, especially for \emph{DeepLab} and \emph{class-aware RPN}. The improvement saturates when using $3$ deformable layers for \emph{DeepLab}, and $6$ for others. In the remaining experiments, we use $3$ in the feature extraction networks.

We empirically observed that the learned offsets in the deformable convolution layers are highly adaptive to the image content, as illustrated in Figure~\ref{fig.two_layer_receptive_field_example} and Figure~\ref{fig.deform_conv_example}. To better understand the mechanism of deformable convolution, we define a metric called \emph{effective dilation} for a deformable convolution filter. It is the mean of the distances between all adjacent pairs of sampling locations in the filter. It is a rough measure of the receptive field size of the filter.

We apply the R-FCN network with $3$ deformable layers (as in Table~\ref{table.ablation_usage_deformable_convolution}) on VOC 2007 test images. We categorize the deformable convolution filters into four classes: small, medium, large, and background, according to the ground truth bounding box annotation and where the filter center is. Table~\ref{table.deformable_convolution_stat} reports the statistics (mean and std) of the effective dilation values. It clearly shows that: 1) \emph{the receptive field sizes of deformable filters are correlated with object sizes, indicating that the deformation is effectively learned from image content}; 2) \emph{the filter sizes on the background region are between those on medium and large objects, indicating that a relatively large receptive field is necessary for recognizing the background regions}. These observations are consistent in different layers.

The default ResNet-101 model uses atrous convolution with dilation $2$ for the last three $3\times 3$ convolutional layers (see Section~\ref{sec.deformable_convnets}). We further tried dilation values $4$, $6$, and $8$ and reported the results in Table~\ref{table.ablation_VOC}. It shows that: 1) \emph{accuracy increases for all tasks when using larger dilation values, indicating that the default networks have too small receptive fields}; 2) \emph{the optimal dilation values vary for different tasks, \eg, $6$ for DeepLab but $4$ for Faster R-CNN;} 3) \emph{deformable convolution has the best accuracy}. These observations verify that adaptive learning of filter deformation is effective and necessary.

\textbf{Deformable RoI Pooling} It is applicable to \emph{Faster R-CNN} and \emph{R-FCN}. As shown in Table~\ref{table.ablation_VOC}, using it alone already produces noticeable performance gains, especially at the strict mAP@0.7 metric. When both deformable convolution and RoI Pooling are used, significant accuracy improvements are obtained.

\textbf{Model Complexity and Runtime} Table~\ref{table.complexity_memory_runtime} reports the model complexity and runtime of the proposed deformable ConvNets and their plain versions. Deformable ConvNets only add small overhead over model parameters and computation. This indicates that the significant performance improvement is from the capability of modeling geometric transformations, other than increasing model parameters.

\subsection{Object Detection on COCO}
\label{sec.experiment_comparison}

In Table~\ref{table.detection_on_coco}, we perform extensive comparison between the deformable ConvNets and the plain ConvNets for object detection on COCO test-dev set. We first experiment using ResNet-101 model.  The deformable versions of class-aware RPN, Faster R-CNN and R-FCN achieve mAP@[0.5:0.95] scores of 25.8\%, 33.1\%, and 34.5\% respectively, which are 11\%, 13\%, and 12\% relatively higher than their plain-ConvNets counterparts respectively. By replacing ResNet-101 by Aligned-Inception-ResNet in Faster R-CNN and R-FCN, their plain-ConvNet baselines both improve  thanks to the more powerful feature representations. And the effective performance gains brought by deformable ConvNets also hold. By further testing on multiple image scales (the image shorter side is in [480, 576, 688, 864, 1200, 1400]) and performing iterative bounding box average~\cite{gidaris2015object}, the mAP@[0.5:0.95] scores are increased to 37.5\% for the deformable version of R-FCN. Note that the performance gain of deformable ConvNets is complementary to these bells and whistles. 

\section{Conclusion}

This paper presents deformable ConvNets, which is a simple, efficient, deep, and end-to-end solution to model dense spatial transformations. For the first time, we show that it is feasible and effective to learn dense spatial transformation in CNNs for sophisticated vision tasks, such as object detection and semantic segmentation.

\subsubsection*{Acknowledgements}
The Aligned-Inception-ResNet model was trained and investigated by Kaiming He, Xiangyu Zhang, Shaoqing Ren, and Jian Sun in unpublished work.

\appendix

\section{Deformable Convolution/RoI Pooling Back-propagation}
\label{sec:deformable_conv_backward}

In the deformable convolution Eq.~\eqref{eq.deformable_conv}, the gradient w.r.t. the offset $\Delta \mathbf{p}_n$ is computed as
\begin{equation}
\begin{split}
\frac{\partial \mathbf{y}(\mathbf{p}_0)}{\partial \Delta \mathbf{p}_n} & = \sum_{\mathbf{p}_n\in\mathcal{R}}\mathbf{w}(\mathbf{p}_n)\cdot \frac{\partial \mathbf{x}(\mathbf{p}_0+\mathbf{p}_n+\Delta \mathbf{p}_n)}{\partial \Delta \mathbf{p}_n} \\
= \sum_{\mathbf{p}_n\in\mathcal{R}} & \left[\mathbf{w}(\mathbf{p}_n)\cdot \sum_\mathbf{q}  \frac{\partial  G(\mathbf{q}, \mathbf{p}_0+\mathbf{p}_n+\Delta \mathbf{p}_n)}{\partial \Delta \mathbf{p}_n} \mathbf{x}(\mathbf{q})\right], \\
\end{split}
\end{equation}

where the term $\frac{\partial  G(\mathbf{q}, \mathbf{p}_0+\mathbf{p}_n+\Delta \mathbf{p}_n)}{\partial \Delta \mathbf{p}_n} $ can be derived from Eq.~\eqref{eq.bilinear_kernel}. Note that the offset $\Delta \mathbf{p}_n$ is 2D and we use $\partial \Delta \mathbf{p}_n$ to denote $\partial\Delta p_n^x$ and $\partial\Delta p_n^y$ for simplicity.

Similarly, in the deformable RoI Pooling module, the gradient w.r.t. the offset $\Delta \mathbf{p}_{ij}$ can be computed by
\begin{equation}
\begin{split}
\frac{\partial \mathbf{y}(i,j)}{\partial \Delta \mathbf{p}_{ij}} & =\frac{1}{n_{ij}} \sum_{\mathbf{p}\in bin(i,j)} \frac{\partial \mathbf{x}(\mathbf{p}_0+\mathbf{p}+\Delta \mathbf{p}_{ij})}{\partial \Delta \mathbf{p}_{ij}} \\
=\frac{1}{n_{ij}} & \sum_{\mathbf{p}\in bin(i,j)} \left[ \sum_\mathbf{q}  \frac{\partial  G(\mathbf{q}, \mathbf{p}_0+\mathbf{p}+\Delta \mathbf{p}_{ij})}{\partial \Delta \mathbf{p}_{ij}} \mathbf{x}(\mathbf{q}) \right]. \\
\end{split}
\end{equation}
And the gradient w.r.t. the normalized offsets $ \Delta \widehat{\mathbf{p}}_{ij}$ can be easily obtained via computing derivatives in $\Delta \mathbf{p}_{ij} = \gamma \cdot \Delta \widehat{\mathbf{p}}_{ij} \circ (w, h)$.

\section{Details of Aligned-Inception-ResNet}
\label{sec:aligned_inception_resnet}

\newcommand{\blockb}[3]{\multirow{3}{*}{
\(\left[
\begin{array}{l}
\text{1$\times$1, #2}\\
[-.1em] \text{3$\times$3, #2}\\
[-.1em] \text{1$\times$1, #1}
\end{array}\right]\)$\times$#3}
}

\newcommand{\blockincres}[2]{\tabincell{c}{\multirow{3}{*}{
\(\left[
\begin{array}{c}
\text{#1-d}\\
[-.1em] \text{IRB}\\
\end{array}\right]\)$\times$#2}
}}

\newcolumntype{x}[1]{>\centering p{#1pt}}
\renewcommand\arraystretch{1.25}
\setlength{\tabcolsep}{8pt}
\begin{table}[t]
\begin{center}
\begin{tabular}{c|c|c}
\hline
 stage & spatial dim. & Aligned-Inception-ResNet \\
\hline \hline
conv1 & 112$\times$112 & 7$\times$7, 64, stride 2 \\
\hline
\multirow{4}{*}{conv2} & \multirow{4}{*}{ 56$\times$56} & 3$\times$3 max pool, stride 2 \\\cline{3-3}
  &   & {\blockincres{256}{3}}\\
  &   & \\
  &   & \\
\hline
\multirow{3}{*}{conv3} &  \multirow{3}{*}{ 28$\times$28}
   &  {\blockincres{512}{4}}\\
  &   & \\
  &  & \\
\hline
\multirow{3}{*}{conv4} & \multirow{3}{*}{ 14$\times$14}
   & \blockincres{1024}{23}\\
  &   & \\
  &   & \\
\hline
\multirow{3}{*}{conv5} & \multirow{3}{*}{ 7$\times$7}
 & \blockincres{2048}{3}\\
  &   & \\
  &   & \\
\hline
\multirow{2}{*}{classifier} & \multirow{2}{*}{ 1$\times$1} & global average pool, \\
 & & 1000-d fc, softmax \\
\hline
\end{tabular}
\end{center}
\caption{Network architecture of Aligned-Inception-ResNet. The \textit{Inception Residual Block} (IRB) is detailed in Figure~\ref{fig.inception_residual_module_v4}.}
\label{tab:architecture}
\vspace{-.5em}
\end{table}

\setlength{\tabcolsep}{4.0pt}
\renewcommand{\arraystretch}{1.2}
\begin{table}[t]
\centering
\small
\begin{tabular}{l|c|c|c}
\hline
Network & \# params & \renewcommand{\arraystretch}{1.0} \tabincell{c}{top-1 err (\%)} & \renewcommand{\arraystretch}{1.0} \tabincell{c}{top-5 err (\%)} \\
\hline
\hline{\scriptsize ResNet-101}  &  $46.0$M & 23.6 & 7.1 \\
{\scriptsize Inception-ResNet-v2}  & $54.3$M & 19.6 & 4.7 \\ 
{\scriptsize Aligned-Inception-ResNet} & $64.3$M & 22.1 & 6.0 \\ 
\hline
\end{tabular}
\caption{Comparison of Aligned-Inception-ResNet with ResNet-101 and Inception-ResNet-v2 on ImageNet-1K validation.}
\label{table.complexity_accuracy}
\end{table}

\begin{figure}
  \centering
  \includegraphics[width=0.95\linewidth]{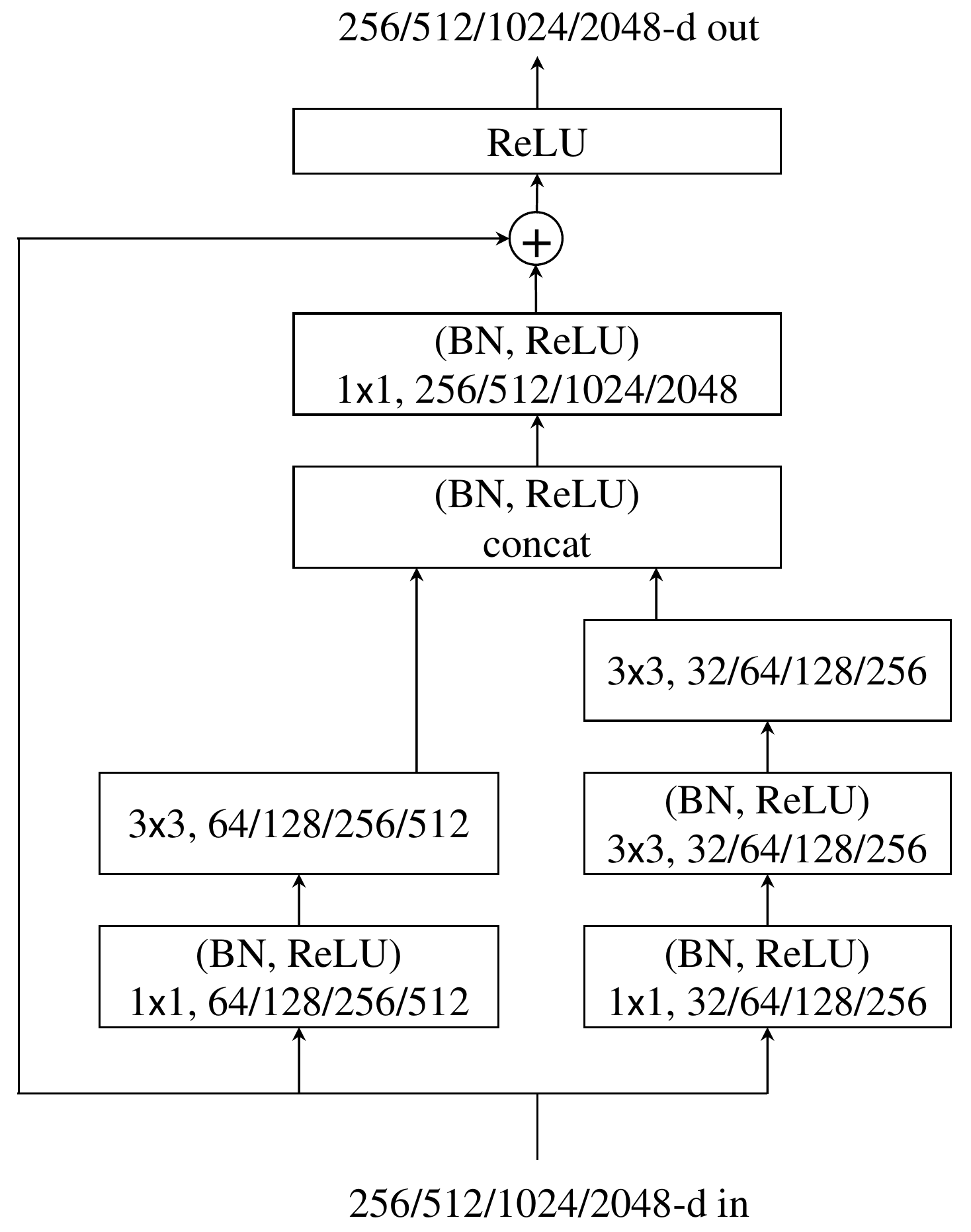}
\caption{The \textit{Inception Residual Block} (IRB) for different stages of Aligned-Inception-ResNet, where the dimensions of different stages are separated by slash (conv2/conv3/conv4/conv5).}
\label{fig.inception_residual_module_v4}
\end{figure}

In the original Inception-ResNet~\cite{szegedy2016inception} architecture, multiple layers of valid convolution/pooling are utilized, which brings feature alignment issues for dense prediction tasks. For a cell on the feature maps close to the output, its projected spatial location on the image is not aligned with the location of its receptive field center. Meanwhile, the task specific networks are usually designed under the alignment assumption. For example, in the prevalent FCNs for semantic segmentation, the features from a cell are leveraged to predict the pixel’s label at the corresponding projected image location.

To remedy this issue, the network architecture is modified~\cite{he2016aligned}, called ``Aligned-Inception-ResNet'' and shown in Table~\ref{tab:architecture}. When the feature dimension changes, a $1\times 1$ convolution layer with stride 2 is utilized. There are two main differences between Aligned-Inception-ResNet and the original Inception-ResNet~\cite{szegedy2016inception}. Firstly, Aligned-Inception-ResNet does not have the feature alignment problem, by proper padding in convolutional and pooling layers. Secondly, Aligned-Inception-ResNet consists of repetitive modules, whose design is simpler than the original Inception-ResNet architectures.

The Aligned-Inception-ResNet model is pre-trained on ImageNet-1K classification~\cite{deng2009imagenet}. The training procedure follows ~\cite{he2016deep}. Table~\ref{table.complexity_accuracy} reports the model complexity, top-1 and top-5 classification errors.

{\small
\bibliographystyle{ieee}
\bibliography{egbib}
}

\end{document}